\documentclass[conference]{IEEEtran}

\makeatletter

\def\ps@IEEEtitlepagestyle{%
  \def\@evenfoot{}%
}

\usepackage{blindtext}
\usepackage{eso-pic}
\IEEEoverridecommandlockouts

\usepackage{cite}
\usepackage{amsmath,amssymb,amsfonts}
\usepackage{algorithmic}
\usepackage{graphicx}
\usepackage{textcomp}
\usepackage{xcolor}
\def\BibTeX{{\rm B\kern-.05em{\sc i\kern-.025em b}\kern-.08em
    T\kern-.1667em\lower.7ex\hbox{E}\kern-.125emX}}
    
\usepackage{eso-pic}
\newcommand\AtPageUpperMyright[1]{\AtPageUpperLeft{%
 \put(\LenToUnit{0.17\paperwidth},\LenToUnit{-2cm}){%
     \parbox{0.9\textwidth}{\raggedleft\fontsize{8}{11}\selectfont #1}}%
 }}%
\newcommand{\conf}[1]{%
\AddToShipoutPictureBG*{%
\AtPageUpperMyright{#1}
}
}    

\begin{document}
\title{\vspace*{1cm} Real-Time On-the-Go Annotation Framework Using YOLO for Automated Dataset Generation\\

\thanks{This research is based upon work supported by North
Dakota State University and the U. S. Department of Agri-
culture, Agricultural Research Service, under agreement No.
58-6064-3-011.}
}

\author{\IEEEauthorblockN{Mohamed Abdallah Salem}
\IEEEauthorblockA{
\textit{dept. Agricultural and Biosystems Engineering} \\
\textit{North Dakota State University}\\
Fargo, USA \\
Mohamed.Salem@ndsu.edu}
\and
\IEEEauthorblockN{Ahmed Harb Rabia}
\IEEEauthorblockA{
\textit{dept. Agricultural and Biosystems Engineering} \\
\textit{North Dakota State University}\\
Fargo, USA \\
Ahmed.Rabia@ndsu.edu}
}

\maketitle
\conf{\textit{ Accepted for publication in the Proceedings of the 5. Interdisciplinary Conference on Electrics and Computer (INTCEC 2025) \\ 
15-16 September 2025, Chicago-USA}}

\begin{abstract}
Efficient and accurate annotation of datasets remains a significant challenge for deploying object detection models such as You Only Look Once (YOLO) in real-world applications, particularly in agriculture where rapid decision-making is critical. Traditional annotation techniques are labor-intensive, requiring extensive manual labeling post data collection. This paper presents a novel real-time annotation approach leveraging YOLO models deployed on edge devices, enabling immediate labeling during image capture. To comprehensively evaluate the efficiency and accuracy of our proposed system, we conducted an extensive comparative analysis using three prominent YOLO architectures (YOLOv5, YOLOv8, YOLOv12) under various configurations: single-class versus multi-class annotation and pretrained versus scratch-based training. Our analysis includes detailed statistical tests and learning dynamics, demonstrating significant advantages of pretrained and single-class configurations in terms of model convergence, performance, and robustness. Results strongly validate the feasibility and effectiveness of our real-time annotation framework, highlighting its capability to drastically reduce dataset preparation time while maintaining high annotation quality.
\end{abstract}


\section{Introduction}
The advent of precision agriculture has underscored the necessity for rapid, accurate, and scalable methods to monitor and manage crop health and weed proliferation. Traditional manual annotation of agricultural datasets is labor-intensive, time-consuming, and prone to human error, thereby impeding the timely deployment of machine learning models in dynamic field conditions \cite{mamat2022advanced}. As the agricultural sector increasingly adopts automation and artificial intelligence (AI) technologies, there is a pressing demand for efficient annotation frameworks that can operate in real-time, facilitating immediate data labeling and model refinement \cite{beck2020embedded}.

Object detection models, particularly those based on the You Only Look Once (YOLO) architecture, have demonstrated remarkable efficacy in various agricultural applications, including pest detection, crop monitoring, and yield estimation \cite{badgujar2024agricultural, khan2025objectdetection}. YOLO's real-time processing capabilities and high accuracy make it an ideal candidate for deployment in agricultural settings where rapid decision-making is crucial \cite{hakim2025yolo}. However, the effectiveness of these models is heavily contingent upon the quality and quantity of annotated data, which remains a significant bottleneck in the development pipeline \cite{zhang2023yoloweeds}

To address this challenge, we propose a novel real-time annotation framework that leverages YOLO models deployed on edge devices to facilitate immediate labeling during image capture. This approach aims to streamline the data annotation process, reduce latency between data collection and model training, and enhance the overall efficiency of deploying AI models in agricultural environments. By integrating real-time detection and annotation, our framework seeks to empower agricultural practitioners with tools that can adapt swiftly to the evolving conditions of the field.

This paper introduces an on-the-go annotation framework that integrates YOLO-based real-time object detection with live manual class labeling and YOLO-format export. Our contributions include:
\begin{itemize}
    \item A lightweight annotation pipeline deployable on edge devices.
    \item A live labeling interface for in-field selective annotation.
    \item Evaluation of 12 YOLO model training configurations.
    \item Statistical analysis comparing single vs. multi-class and pretrained vs. scratch training.
\end{itemize}

\begin{figure*}[ht]
    \centering
    \includegraphics[width=\textwidth]{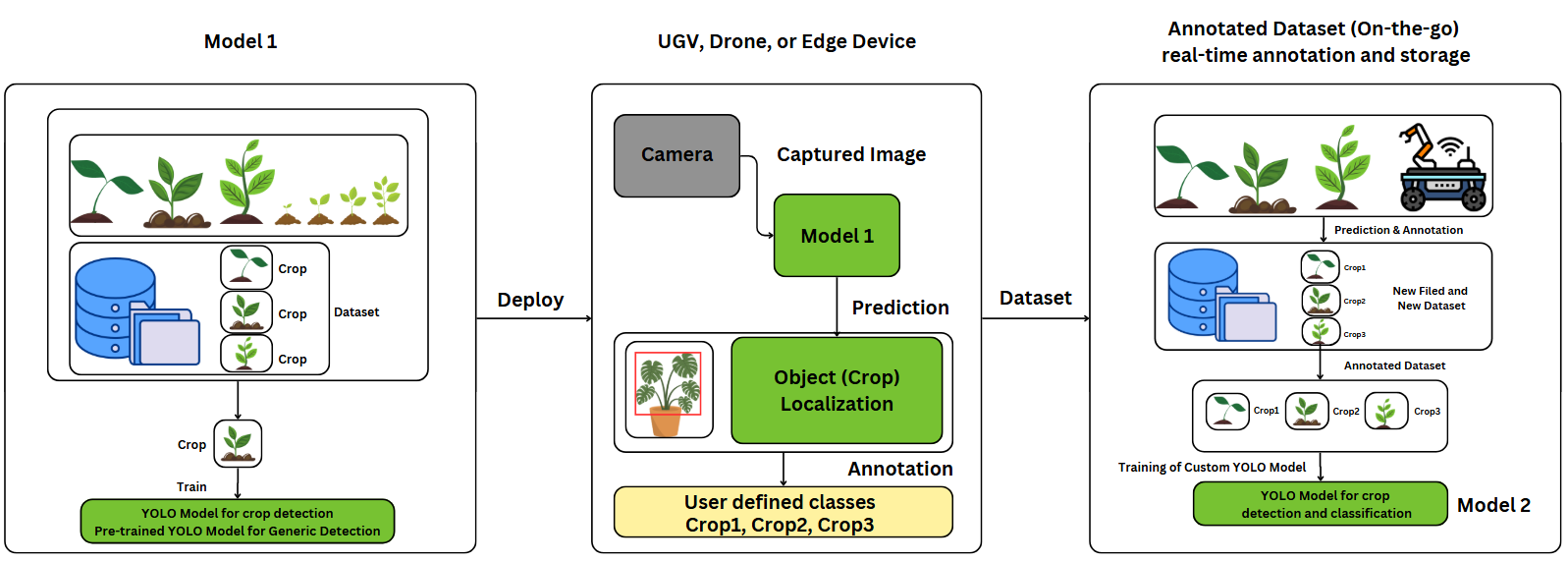}
    \caption{Overview of the proposed real-time YOLO annotation framework.}
    \label{fig:annotation_framework}
\end{figure*}

\section{Related Work}

The integration of deep learning techniques in agriculture has witnessed substantial growth, with object detection models like YOLO playing a pivotal role in advancing precision farming practices. YOLO's ability to perform real-time object detection has been harnessed in various agricultural applications, such as identifying pests on crops, monitoring plant health, and facilitating automated harvesting \cite{badgujar2024agricultural, dalal2025systematic, agripest2023, pestlite2023}. These applications underscore the model's versatility and its potential to revolutionize traditional farming methodologies.

Recent studies have focused on enhancing YOLO's performance in agricultural contexts. For instance, the development of lightweight YOLO variants, such as YOLO-YSTs, has aimed to balance detection accuracy with computational efficiency, enabling deployment on edge devices for tasks like pest detection on yellow sticky traps \cite{huang2025yolo}. Similarly, models like Ag-YOLO have been tailored for resource-constrained hardware, achieving real-time detection speeds while maintaining high accuracy in identifying agricultural objects \cite{zhang2021ag}.

Despite these advancements, the annotation process remains a critical hurdle. Traditional annotation methods are often disconnected from the data collection phase, leading to delays and potential inconsistencies in labeling \cite{rasmussen2022challenge}. Efforts to automate annotation using AI models have shown promise, yet they frequently require substantial post-processing and validation to ensure accuracy \cite{wang2025dynamic, nong2022semi}. The need for integrated systems that can perform detection and annotation simultaneously in real-time is evident, particularly in scenarios where rapid response is essential \cite{bao2025zero}.

Our proposed framework builds upon these developments by introducing a real-time annotation system that operates concurrently with data acquisition. By deploying YOLO models on edge devices, we enable immediate detection and labeling of agricultural objects during image capture. This integration aims to reduce the annotation burden, accelerate the training cycle of AI models, and enhance the responsiveness of agricultural monitoring systems. Through this approach, we seek to contribute a practical solution to the ongoing challenge of efficient data annotation in precision agriculture.

\section{Methodology}
\subsection{Experimental Design and Workflow Overview}

The primary aim of this study was to develop and validate an efficient, real-time annotation framework for YOLO-based object detection models, specifically tailored for agricultural applications such as weed and crop detection. To comprehensively evaluate this framework, we designed systematic experiments involving several key components: data selection and preparation, model training strategies, model deployment for real-time annotation, and robust statistical validation. The study was conducted through a clearly structured workflow divided into sequential and interconnected stages, ensuring clarity, reproducibility, and thorough evaluation of results.

\subsection{Dataset Description and Preparation}

A crucial component of the experimental methodology was the selection of an appropriate dataset. We utilized a comprehensive, publicly available agricultural dataset composed of RGB images representing a diverse range of species called Weed-crop dataset \cite{upadhyay2025weed}. Specifically, the dataset included images of eight crop species---Black Bean, Canola, Corn, Lentil, Field Pea, Flax, Soybean, and Sugar Beet---and five weed species---Kochia, Horseweed, Water Hemp, Ragweed, and Redroot Pigweed. The dataset's rich diversity in species facilitated robust training and evaluation of the YOLO architectures across different classification complexities.

Each image in the dataset was accompanied by precise annotations in YOLO format, specifying object class labels and bounding boxes. To explore the impacts of class granularity on detection accuracy, the dataset underwent two separate preparation procedures. In the first preparation stage, the dataset retained its original multi-class labeling structure (13 classes: 8 crops, 5 weeds). In the second preparation stage, annotations were generalized into a single \texttt{Plant} class to simplify the object detection for real-time annotation task and make it general purpose, enabling the user to add new classes based on single-class annotation strategies. Both datasets were subsequently partitioned into standard training (70\%), validation (15\%), and test (15\%) subsets, maintaining balanced representation of all classes to avoid bias. 

To enhance the robustness and generalization capabilities of the YOLO models under diverse field conditions, a comprehensive preprocessing and augmentation pipeline was applied prior to training. Images were first auto-oriented and resized using a stretched scaling method to fit a uniform resolution of 640X640 pixels, ensuring compatibility with the input requirements of all YOLO versions. Each training image underwent augmentation to generate three distinct variants, resulting in a total of 2,549 images used during training.

Augmentation transformations included horizontal and vertical flipping, as well as $90^{\circ}$ rotations applied in clockwise, counter-clockwise, and upside-down orientations. Additionally, color-based augmentations were introduced to simulate variable lighting conditions and sensor exposure differences commonly encountered in field deployments. Specifically, saturation was adjusted randomly within a range of $\pm 25\%$, brightness within $\pm 15\%$, and exposure within $\pm 10\%$. These transformations collectively aimed to increase model resilience to real-world imaging variability, such as changes in ambient light, orientation of plant structures, and background clutter.

By incorporating this diverse set of augmentations, the training process was optimized to improve the model's ability to generalize across unseen conditions, reducing the likelihood of overfitting and improving the reliability of real-time detections during deployment.

\subsection{YOLO Model Architectures and Configurations}

Our experimental framework encompassed the evaluation of three popular YOLO architectures: YOLOv5, YOLOv8, and the recently introduced YOLOv12. Each model was assessed comprehensively under four distinct training conditions, yielding twelve experimental scenarios. These training conditions included combinations of training from scratch (no pretrained weights) versus employing pretrained weights derived from COCO datasets, as well as comparisons between single-class and multi-class annotations.

For each training scenario, hyperparameter optimization involved systematic adjustments of learning rate (initial learning rate: $1 \times 10^{-3}$), batch size (set to 32), and number of epochs (fixed at 100), selected through preliminary trials to balance computational cost and performance.

\subsection{Training and Evaluation Procedure}

All model training and evaluations were conducted on a Lambda Vector workstation equipped with dual NVIDIA RTX 4500 Ada Generation GPUs, each with 24 GB of dedicated memory. The system utilized NVIDIA Driver version 550.144.03 and CUDA 12.4, running a PyTorch-based training environment. We employed the official Ultralytics implementations for YOLOv5, YOLOv8, and YOLOv12, ensuring consistency across experiments and full compatibility with model architecture updates. GPU acceleration enabled efficient batch processing and allowed detailed tracking of training progression through metrics such as mean Average Precision (mAP@\{50--95\}), precision, recall, and F1-score. These metrics were selected to reflect both the detection accuracy and the operational viability of models in real-world agricultural deployments.

Model training involved continuous performance monitoring to avoid overfitting and ensure optimal convergence. The validation set was employed during training epochs to perform checkpointing, ensuring that the final evaluated models represented the highest achievable performance within each experimental configuration. Upon completion, performance metrics were extracted and consolidated for comprehensive statistical analysis.

\subsection{Real-Time Annotation Framework Implementation}

\begin{figure}[ht]
    \centering
    \includegraphics[width=\columnwidth]{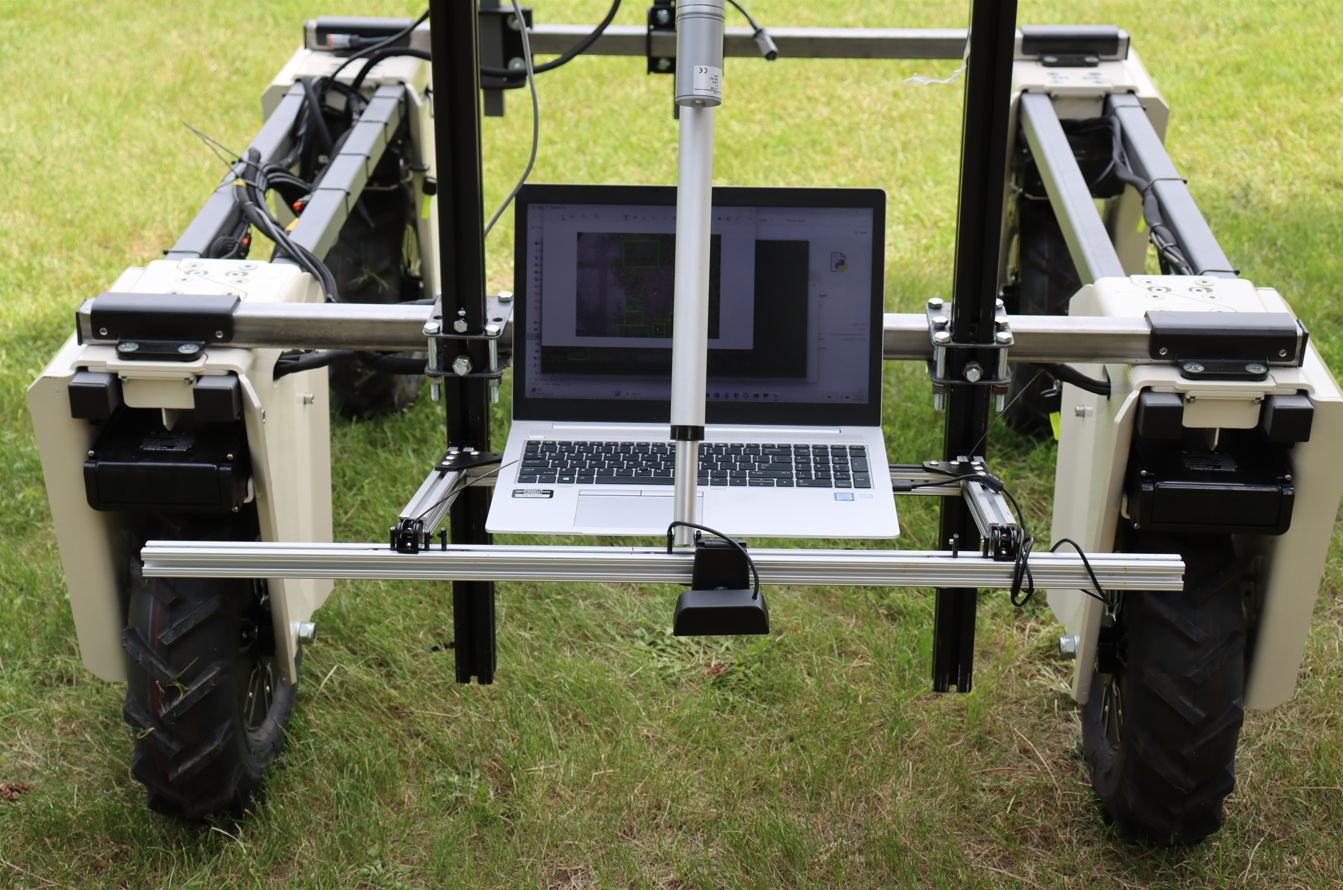}
    \caption{Field deployment setup for real-time image capture and annotation using a mounted camera and edge computing system on an unmanned ground vehicle (UGV). The laptop runs the YOLO model inference interface and displays live predictions.}
    \label{fig:ugv-setup}
\end{figure}

A critical methodological innovation was the integration of YOLO models into a real-time, on-the-go annotation system. A specialized software interface was developed in Python, leveraging OpenCV for real-time camera image acquisition and model inference. This interface enabled immediate visualization of detected objects via bounding boxes with confidence scores. Users could confirm or adjust predictions in real-time by pressing specific keys (e.g., the \texttt{Enter} key to save and annotate the image). This setup was deployed on an edge device (a laptop or embedded system) mounted directly on an unmanned ground vehicle (UGV) as shown in Fig.~\ref{fig:ugv-setup}, facilitating practical field-based data acquisition.

During field tests, real-time detections were displayed continuously, providing immediate feedback regarding model performance and annotation quality. Captured images and corresponding YOLO-formatted annotation files were stored directly on the device, enabling instantaneous dataset expansion. This real-time system was validated in diverse field conditions, including varying lighting, background complexity, and object densities, replicating realistic scenarios likely encountered by agricultural practitioners.

\subsection{Statistical Validation}

To substantiate observed differences across model configurations rigorously, robust statistical analyses were performed. Independent two-sample t-tests evaluated the statistical significance of observed performance variations between different training methods (single-class vs. multi-class, pretrained vs. scratch). Specific hypotheses tested included whether pretrained models significantly outperform scratch-trained models and whether single-class configurations effectively simplify annotation without significant loss of performance.

All statistical tests utilized a significance threshold of $\alpha = 0.05$, ensuring rigorous evaluation standards. Test outcomes were systematically reported in terms of t-statistics and corresponding p-values. Results demonstrating p-values below this threshold were considered statistically significant, validating differences in performance metrics across configurations.

\begin{figure*}[ht]
    \centering
    \includegraphics[width=\textwidth]{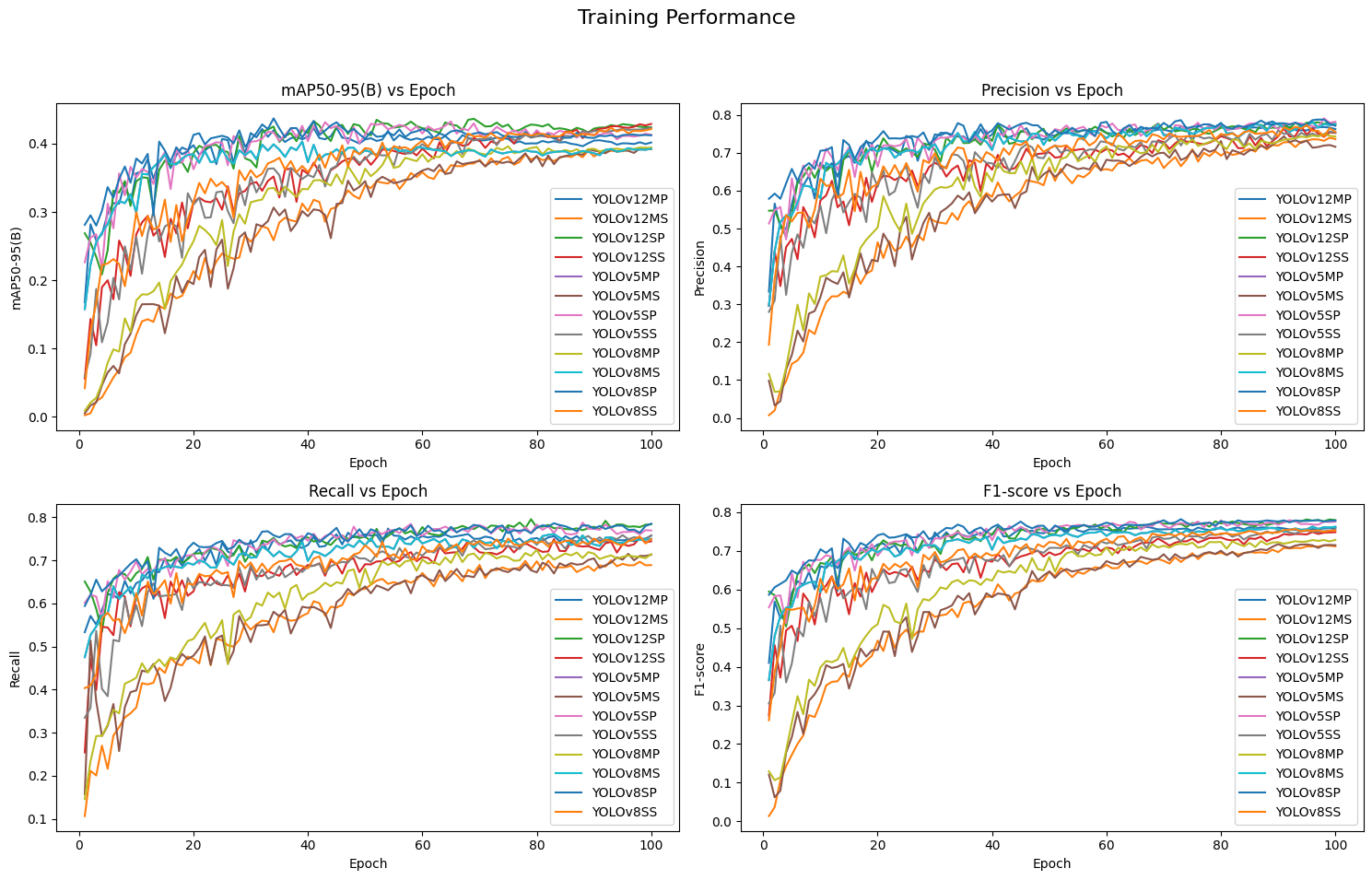}
    \caption{Performance metrics across 100 epochs for YOLO models under various configurations.}
    \label{fig:training_performance}
\end{figure*}

\subsection{Data Visualization and Interpretation}

Results were comprehensively visualized using various statistical and graphical techniques. Box plots illustrated distributions, variances, and median values across different model configurations, providing clear comparative insights. Performance metrics evolution over training epochs was visualized through line plots, offering intuitive assessments of convergence dynamics and model stability over time. Furthermore, histograms of mAP@\{50--95\} distributions provided concise visual summaries of overall model reliability, indicating performance consistency across scenarios.

\section{Experimental Results}
Extensive experiments were conducted using YOLOv5, YOLOv8, and YOLOv12 under single-class and multi-class labeling schemes with pretrained and scratch-based configurations. The results highlight pretrained configurations' significant superiority across key metrics such as mean average precision (mAP), precision, recall, and F1-score.

\begin{figure}[ht]
    \centering
    \includegraphics[width=\columnwidth]{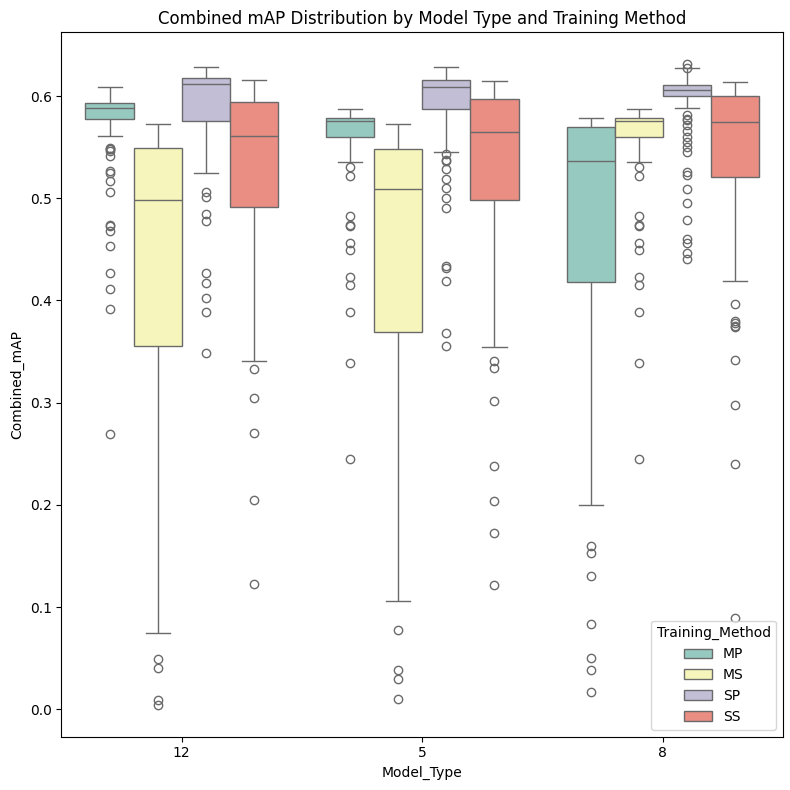}
    \caption{Comparative performance by YOLO model type and training method.}
    \label{fig:model_comparison}
\end{figure}

\subsection{Training Performance and Learning Dynamics}
Analysis of training curves across epochs (as depicted clearly in Fig.~\ref{fig:training_performance}) reveals distinct learning behaviors between the pretrained and scratch-based models. The pretrained configurations --- particularly YOLOv12MP, YOLOv12SP, YOLOv8MP, and YOLOv8SP --- demonstrate rapid convergence, attaining higher mAP@50--95 scores early in training, often within the first 30 epochs. This rapid convergence can be attributed to the initialization of models with weights pretrained on extensive generic datasets, enabling more efficient and accurate transfer learning even when applied to specialized agricultural tasks such as weed detection and crop classification. Conversely, models trained from scratch (e.g., YOLOv5MS, YOLOv8MS, and YOLOv12MS) showed significantly slower convergence and higher variability in training trajectories, indicating challenges in optimizing weights from random initialization, particularly given limited training data for specialized applications.

\begin{figure}[ht]
    \centering
    \includegraphics[width=0.48\textwidth]{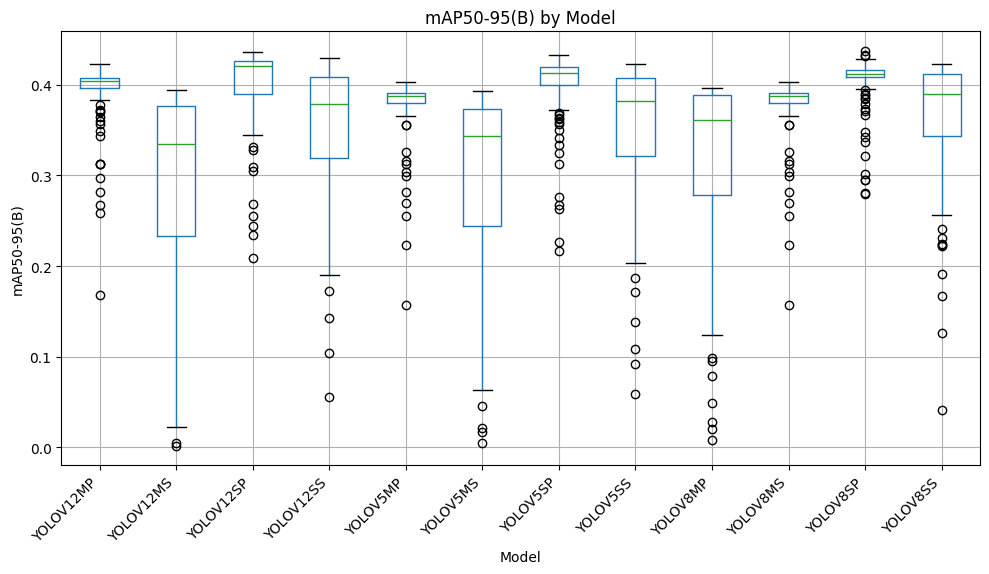}
    \caption{Detailed mAP@50-95 comparison across models.}
    \label{fig:detailed_boxplot}
\end{figure}

Moreover, precision, recall, and F1-scores followed similar patterns. Pretrained single-class (SP) and multi-class (MP) configurations consistently yielded higher precision and recall, suggesting superior ability in accurately identifying relevant classes while maintaining fewer false positives. Notably, pretrained single-class configurations exhibited an advantage in scenarios focused solely on detecting the presence of any plant class, without specific subclass distinction, demonstrating the effective applicability of simplified annotation strategies in real-time contexts. On the other hand, multi-class pretrained models (MP) excelled in scenarios demanding finer discrimination between closely related species, indicated by higher recall and balanced F1-scores towards the later epochs of training.

\subsection{Comparative Performance Analysis by Model Type and Training Method}
The box plots presented in Fig.~\ref{fig:model_comparison} provide an insightful quantitative comparison across the different YOLO configurations. The results distinctly indicate that pretrained configurations significantly outperform scratch-based training methods. Specifically, YOLOv12MP and YOLOv12SP exhibit higher median mAP@50-95 and reduced interquartile ranges, implying not only better average performance but also enhanced reliability and consistency across different dataset splits or subsets. Conversely, scratch-based multi-class models (e.g., YOLOv5MS and YOLOv8MS) showed the greatest variability and generally lower median performance. Such variability may originate from random initializations leading to local minima during training or difficulty in adequately generalizing from limited domain-specific datasets.

Detailed examination in Fig.~\ref{fig:detailed_boxplot} further confirms these observations. YOLOv12-based models dominate with consistently higher medians and reduced variance compared to YOLOv5 and YOLOv8 models, regardless of training configurations. YOLOv8 models follow closely behind, with pretrained setups consistently outperforming from-scratch variants. The observed performance hierarchy-YOLOv12 $>$ YOLOv8 $>$ YOLOv5-highlights architectural improvements and optimization strategies introduced in later YOLO iterations, reinforcing their applicability and robustness in practical field applications involving challenging real-world conditions, such as varying illumination and complex backgrounds.

\subsection{Statistical Significance and Validation}

To statistically validate the observations, independent t-tests were conducted, yielding results that strongly support the observed empirical differences. For instance, comparing YOLOv5 single-class pretrained (SP) and multi-class pretrained (MP) models, the significant t-test result (T-statistic = 2.5588, P-value = 0.0113) confirms that single-class pretrained training leads to significantly better F1-scores, underscoring its effectiveness for scenarios with simplified detection requirements. Similarly, YOLOv12 single-class scratch (SS) and multi-class scratch (MS) configurations demonstrate a highly significant difference in recall (T-statistic = -5.5068, P-value $<$ 0.0001), suggesting multi-class labeling strategies are crucial for maximizing recall when training from scratch, likely due to increased contextual information available from multiple classes.

Additionally, YOLOv8 single class scratch versus pretrained (T-statistic = -5.1995, P-value $<$ 0.0001) comparisons clearly reveal that pretrained models substantially outperform their scratch-trained counterparts, reinforcing the recommendation to adopt pretrained strategies whenever possible to leverage existing generalized features from large-scale datasets.

\subsection{Implications of Single-Class Training for On-the-Go Annotation}

Beyond performance metrics, one of the key contributions of this work is the strategic use of single-class YOLO models to support real-time, on-the-go data annotation workflows. In field deployments where the objective is not immediate multi-class classification but rather \textit{rapid bounding box generation for any relevant object}, single-class training offers a practical advantage. These models are optimized to detect a general category-such as vegetation, weed presence, or crop material-regardless of specific class labels. This allows users to interactively assign labels post-prediction or during capture, significantly streamlining the annotation process.

The statistical superiority of single-class pretrained models in F1-score (e.g., YOLOv5 SP vs. MP, $p = 0.0113$) supports this operational philosophy: they are more precise in localization and confident detection even when class granularity is reduced. This is ideal for bootstrapping new datasets in underrepresented domains where class diversity evolves during the data collection process.

Moreover, using a single detection class reduces computational complexity, enabling smoother real-time inference on edge devices. It also mitigates model confusion in noisy field conditions where visual overlap between similar species is common. This lightweight, class-agnostic detection pipeline enhances model generalizability, makes labeling more user-driven, and accelerates the training cycle by removing the need for large, class-balanced datasets at early stages.

In this context, the single-class configuration is not a limitation but a deliberate design choice that aligns with the practical demands of field-level annotation and iterative model refinement in precision agriculture.

\subsection{Inference Time Analysis}

\begin{figure}[ht]
    \centering
    \includegraphics[width=\columnwidth]{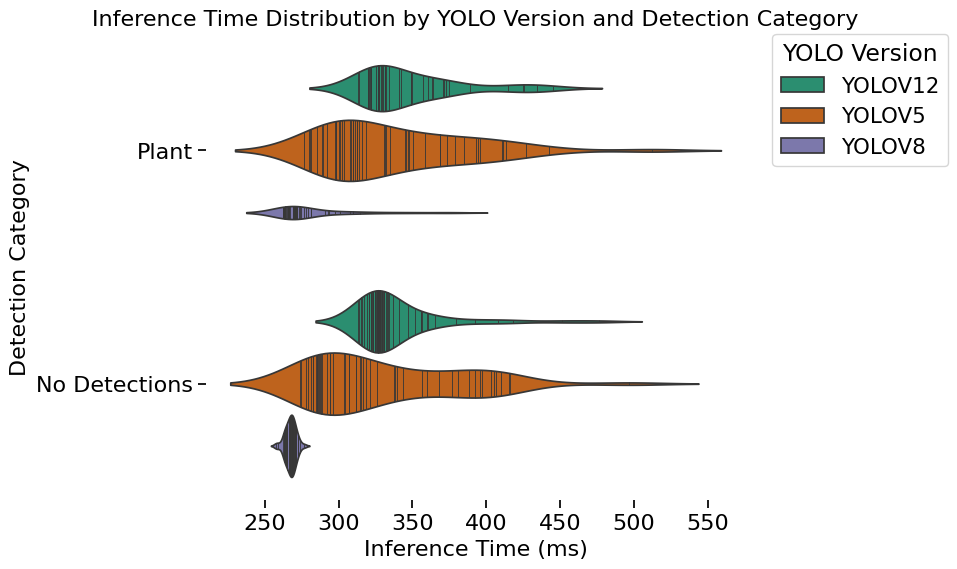}
    \caption{Inference time distribution of different models across plant detections and non-detections.}
    \label{fig:v5}
\end{figure}

In addition to evaluating detection accuracy, precision, and recall, we examined the real-time inference capabilities of the deployed models by analyzing their frame-wise processing latency under actual deployment conditions. Figure~\ref{fig:v5} illustrate the inference time distributions for YOLOv5, YOLOv8, and YOLOv12, segregated by detection outcomes (i.e., frames with and without plant detection). The violin plots capture the distributional characteristics of processing time, offering insight into consistency and responsiveness.

Among the three models, YOLOv8 achieved the fastest average inference time, registering a mean of 275.71 ms per frame. This efficiency is especially critical for time-sensitive applications in precision agriculture, such as robotic navigation and selective spraying, where minimal latency is essential. YOLOv5 followed closely with a mean inference time of 335.02 ms, balancing moderate computational load with high detection reliability. YOLOv12, while yielding strong accuracy metrics, exhibited the highest average inference time at 344.12 ms, indicating a trade-off between accuracy and speed.

Across all models, a consistent trend was observed: frames containing detectable plant instances exhibited marginally higher inference latency compared to those without detections. This can be attributed to the additional post-processing time required for bounding box predictions and class confidence computation. Notably, YOLOv8 maintained both high consistency and low variance across detection and non-detection conditions, reinforcing its suitability for embedded deployment where deterministic processing speed is advantageous.

Overall, the inference time analysis underscores the operational readiness of the proposed system, with all models maintaining sub-500 ms latency, enabling near real-time feedback loops in autonomous platforms. These findings support the deployment of YOLOv8 in edge-AI setups for scalable agricultural monitoring, while highlighting YOLOv5 and YOLOv12 as viable alternatives in scenarios where slightly extended processing windows are acceptable in exchange for model flexibility or robustness.

\subsection{Real-Time Field Evaluation}
\begin{figure}[ht]
    \centering
    \includegraphics[width=\columnwidth]{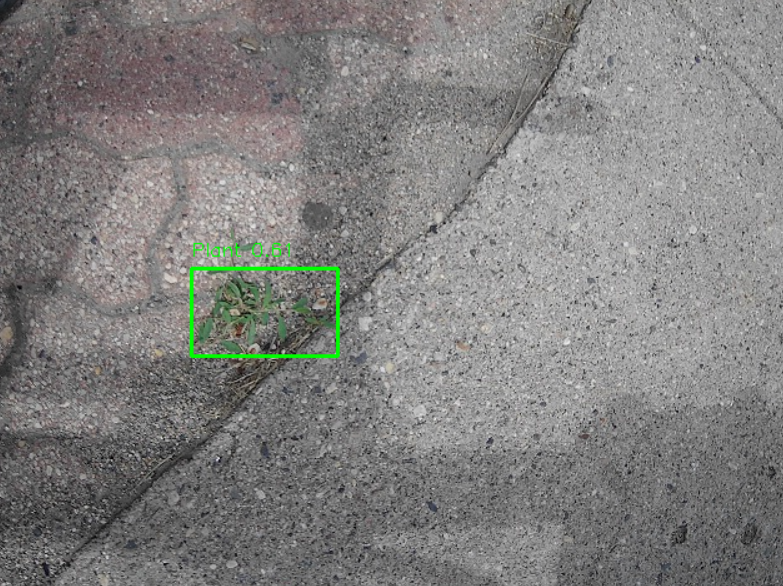}
    \caption{Example detection for grass and random samples that are not exist in the original dataset}
    \label{fig:realtime_examples1}
\end{figure}

Lastly, real-time field deployment results further complement quantitative performance metrics, providing practical validation of the trained models. The inference images (illustrated in Fig.~\ref{fig:realtime_examples1}) demonstrate effective detection capabilities, accurately identifying plants in complex and challenging real-world backgrounds such as concrete pavements with sparse vegetation. These practical outcomes reinforce the feasibility and applicability of deploying pretrained YOLO models in edge computing setups, providing real-time detection and annotation capabilities, thereby significantly streamlining the data annotation process directly in-field.

The comprehensive evaluation and statistical validation clearly articulate several key insights: pretrained configurations significantly enhance model accuracy, consistency, and reliability; single-class annotation simplifies the annotation process without substantial performance sacrifices; and multi-class labeling becomes necessary for maximum recall when training from scratch. Collectively, these findings offer valuable guidance for future deployment strategies of YOLO-based object detection systems, particularly in real-time agricultural contexts where swift annotation and high accuracy are paramount.

\section{Conclusion}
This study introduces a novel real-time annotation framework for precision agriculture, seamlessly integrating image acquisition and labeling through YOLO-based object detection models deployed on edge devices. Unlike traditional workflows that decouple data collection and annotation, our system performs on-the-go labeling during image capture, significantly reducing the time, effort, and latency associated with dataset generation.

We conducted an extensive evaluation across 12 training configurations using YOLOv5, YOLOv8, and YOLOv12, trained on a 13-class dataset comprising crops and weeds. Our results reveal that single-class models-particularly those pretrained-consistently achieved competitive or superior performance compared to multi-class counterparts, not only in real-time inference speed but also across core metrics such as mAP@50-95 and F1-score. These findings are especially notable in the context of precision agriculture, where simplicity, responsiveness, and reliability are paramount.

Statistical t-tests further validated these observations, confirming that single-class training offers measurable advantages in terms of precision and F1-score, while minimizing annotation complexity. This suggests that a general-purpose single-class model can be an effective and efficient strategy for initial deployment or rapid data collection, especially in scenarios where exhaustive multi-class labeling is not feasible.

Live deployment trials confirmed the practical utility of the framework, successfully detecting both seen and unseen plants in uncontrolled outdoor conditions. This real-world generalization underscores the robustness and field-readiness of the proposed solution.

By enabling scalable, efficient, and accurate annotation in dynamic agricultural settings, our work bridges the critical gap between field data collection and model readiness. Future enhancements will focus on integrating multi-modal sensors, implementing voice-guided annotation tools, and developing automated fine-tuning modules directly on edge hardware for continuous learning in the field.

\section*{Acknowledgment}

This research is based upon work supported by North Dakota State University and the U. S. Department of Agriculture, Agricultural Research Service, under agreement No. 58-6064-3-011.

\bibliographystyle{IEEEtran}
\bibliography{references}

\vspace{12pt}
\color{red}

\end{document}